\documentclass[letterpaper]{article} 
\usepackage{aaai24}  
\usepackage{times}  
\usepackage{helvet}  
\usepackage{courier}  
\usepackage[hyphens]{url}  
\usepackage{graphicx} 
\usepackage{natbib}  
\usepackage{caption} 
\usepackage{booktabs}
\usepackage{bbding}
\usepackage{newfloat}
\usepackage[T1]{fontenc}
\usepackage[utf8]{inputenc}
\usepackage{makecell}
\usepackage{multicol}
\usepackage{multirow}
\usepackage{microtype}
\usepackage{newtxmath}
\usepackage{pgfplots}
\usepackage{subcaption}
\usepackage{tablefootnote}
\usepackage{algorithm}
\usepackage{algpseudocode}

\usepackage{listings}
\DeclareCaptionStyle{ruled}{labelfont=normalfont,labelsep=colon,strut=off} 
\floatstyle{ruled}
\newfloat{listing}{tb}{lst}{}
\floatname{listing}{Listing}
\pgfplotsset{compat=1.8}

\title{Advancing Adversarial Robustness Through Adversarial Logit Update}

\author{
    Hao Xuan\textsuperscript{\rm 1},  Peican Zhu\textsuperscript{\rm 2},  Xingyu Li\textsuperscript{\rm 1}
}

\affiliations{
    \textsuperscript{\rm 1} University of Alberta\\
    \textsuperscript{\rm 2} Northwestern Polytechnical University\\
    \{hxuan, xingyu\}@ualberta.ca, ericcan@nwpu.edu.cn
}

\begin{document}

\maketitle

\begin{abstract}

Deep Neural Networks are susceptible to adversarial perturbations. Adversarial training and adversarial purification are among the most widely recognized defense strategies. Although these methods have different underlying logic, both rely on absolute logit values to generate label predictions. In this study, we theoretically analyze the logit difference around successful adversarial attacks from a theoretical point of view and propose a new principle, namely \textbf{\underline{A}}dversarial \textbf{\underline{L}}ogit \textbf{\underline{U}}pdate (ALU), to infer adversarial sample's labels. Based on ALU, we introduce a new classification paradigm that utilizes pre- and post-purification logit differences for model's adversarial robustness boost. Without requiring adversarial or additional data for model training, our clean data synthesis model can be easily applied to various pre-trained models for both adversarial sample detection and ALU-based data classification. Extensive experiments on both CIFAR-10, CIFAR-100, and tiny-ImageNet datasets show that even with simple components, the proposed solution achieves superior robustness performance compared to state-of-the-art methods against a wide range of adversarial attacks. Our Python implementation is submitted in our Supplementary document and will be published upon the paper's acceptance.
\end{abstract}

\section{Introduction}
Deep neural networks (DNNs) show tremendous success in computer vision \citep{ref10,ref12} and natural language processing \citep{ref11}. However, they are vulnerable to adversarial attacks.
Prior arts provide a theoretical framework to explain this observation and conclude that this problem is directly attributed to non-robust features learned in a model \cite{advFeature, advrobustPrior}.
When casting the two general adversarial defense strategies, i.e. adversarial training \citep{ref1} and adversarial purification \citep{ref15}, into the theoretic framework, we find that the former encourages a model to learn more robust features and the latter manipulates data to remedy the negative impact of non-robust features in a pre-trained model after adversarial attacks.

This study aligns with the second strategy and aims to improve the adversarial robustness of a pre-trained model. Ideally, an adversarial purifier is able to remove adversarial perturbations, minimizing the negative impacts of non-robustness features on a pre-trained model's decision. However, since adversarial noise varies with pre-trained models, adversarial attack methods, and data, even if it is not impossible, designing a purifier that can cover all adversarial attacks is extremely challenging. This explains why many state-of-the-art (SOTA) adversarial purification methods show poor performance against different attacks. In this study, we specifically focus on those adversarial data that are misclassified even after adversarial purification and investigate if there is a solution to make their label prediction right. We notice that though one can access both pre- and post-purification logits, i.e. the logits before and after adversarial purification respectively, prior arts always generate the classification results only relying on the post-purification logits. So we arise the following questions:
\begin{center}
\textit{Is it the only method to determine the classification results based on post-purification logits? \\ If not, what other metrics can be used for the decision? }
\end{center}

To answer these questions, we follow the theoretical framework in \cite{advFeature, advrobustPrior} to further analyze the role of robust features and non-robust features in adversarial attacks, and conclude 3 different types of adversarial attacks according to their logit behaviors. Accordingly, we design a plug-in algorithm called Adversarial Logit Update (ALU) to enhance the classification of adversarial samples, even in cases where their post-purification logits yield incorrect predictions. Specifically, unlike prior arts that rely solely on post-purification logits to generate final predictions, ALU fully leverages the valuable information inherent in pre- and post-purification logits, utilizing their difference for decision-making for adversarial samples. To better adapt the ALU algorithm to different adversarial noise, we incorporate test-time adaptation \citep{ref8} in our clean image synthesis model. In addition to boosting the adversarial robustness of various pre-trained models, we also show that ALU can be easily adapted for model training from scratch. Experiments conducted on CIFAR10, CIFAR100 and tiny-ImageNet demonstrate that the solution incorporating ALU significantly enhances model robustness when connected to pre-trained classifiers. Additionally, the unified ALU classifier achieves exceptional performance, surpassing major adversarial training and purification approaches, even without the inclusion of adversarial data during training. 

Our contributions in this study are summarized as follows:
\begin{itemize}
    \item We explore adversarial robustness from a new angle: inferring true labels of adversarial samples by comparing pre- and post-purification logits and propose a new classification paradigm to improve pre-trained models' robustness.
    \item We introduce an ALU-based model-agnostic solution for robust data classification. By purifying adversarial perturbations iteratively in inference, we effectively remedy the negative effect of non-robustness features on label prediction.
    \item We demonstrate that the proposed ALU principle and corresponding method can be readily adapted to various pre-trained classifiers, resulting in a significant performance boost. Moreover, the unified classifier trained with ALU achieves even better robustness, outperforming SOTA methods.
\end{itemize} 

\section{Related Works}
\textbf{Adversarial Attack} is a malicious attack that applies carefully-designed, human imperceptible permutations to the normal data to fool a model. There have been a number of methods proposed to generate adversarial perturbation $\delta$, such as Fast Gradient Sign Method (FGSM) \citep{ref16}, Projected Gradient Descent (PGD) \citep{ref1,ref17}, CW attack~\citep{carlini2017towards}, MIM~\citep{dong2018boosting}, FAB~\citep{croce2020minimally}, and AutoAttack (AA)~\citep{croce2020reliable}, etc. 

\textbf{Adversarial Training} uses a regularization strategy that uses adversarial data as data augmentation in model training to directly enhance model robustness \citep{ref1}. Such a strategy is often referred to as a "min-max" game.

Based on Madry's "min-max" optimization, more advanced adversarial training algorithms are proposed, by incorporating curriculum learning \citep{ref3,ref4,ref18,ref19,Hou2022}, and introducing alternative loss functions \citep{ref2,ref5,ref20, Hou2023}.

\textbf{Adversarial Purification} is another common approach to combat adversarial attacks. It targets to remove adversarial noises from the attacked examples before feeding them into a task-specific model. Usually, adversarial purification can be used as a plug-in module and is applicable to most pre-trained models. Early studies adopt random augmentation, such as cropping, compression and blurring, to distort adversarial perturbation for model robustness. Recently, more sophisticated learning-based methods are proposed. \cite{ref37} uses Denoising Score-Matching (DSM) along with an Energy-Based Model (EBM) to learn a score function that helps denoising the attacked samples. \cite{ref9} takes advantage of the nature of diffusion models to purify adversarial perturbations. It has been shown that adding random Gaussian noise is effective against adversarial perturbations as it "washed out" the well-designed adversarial perturbation \citep{ref39}.
Similarly, \cite{ref28, ref29, ref32} design small noises that perturb the input against the adversarial direction to purify input data. Different from all the other methods, \cite{ref31} utilizes the relation between RGB and RAW images and proposes the use of a learned camera image signal processing (ISP) pipeline to eliminate adversarial noises. 

\textbf{Test-time adaptation} is a technique that explores testing data in auxiliary unsupervised tasks for further model fine-tuning during inference \citep{ref8}. It originally emerges to solve the distribution drift problem in machine learning, but soon finds useful to improve adversarial robustness \citep{ref21}. For example, in adversarial purification, rather than one-step purification, the input data can be updated iteratively during inference \citep{ref28,ref29,ref32}. In addition, test-time adaptation is also exploited to modify either model parameters or activation functions for adversarial robustness in inference. The auxiliary unsupervised tasks for model adversarial robustness improvement include rotation prediction \citep{ref8}, entropy minimization \citep{ref22}, label consistency \citep{ref23}, and test-time contrastive learning \citep{ref24}.

\section{Preliminaries}
\subsection{Problem Setup} 

We consider multi-category classification, where paired training data $\{\mathscr{X},\mathscr{Y}\}=\{(x,y)|x\in \mathbb{R}^{H\times L \times N}, y\in \mathbb{R}^{1\times M}\}$ are drawn from a data distribution $\mathcal{D}$. Here, $H, L, N$ are the dimension of the input $x$, $M$ is the number of categories, $y$ is a one-hot vector indicating the class of the input $x$. A classifier, $\mathcal{C}:\mathscr{X}\xrightarrow{}\mathscr{Y}$, is a function predicting the label $y$ for a given data $x$. That is, $C(x)=y$ for in-distribution data.

Adversarial attack is a malicious attack that applies carefully-designed, human imperceptible permutations $\delta$ to the normal data $x$ to fool a pre-trained classifier. The resulting data $x' = x + \delta$ is called an adversarial sample and $\mathcal{C}(x+\delta)\neq y$. In this paper, we use $'$ to denote variables associated with adversarial samples. 

The goal in this study is to learn a classification model $C$ which is robust against such adversarial perturbations.

\subsection{Feature Decomposition}

In the canonical classification setting, a neural network classifier, $\mathcal{C}=(f_{\theta},W)$, is usually composed of a feature extractor $f_{\theta}$ parameterized by $\theta$ and a weight matrix $W$. $f_{\theta}$ is a function mapping the input $x$ to a real-valued vector $f_{\theta}(x)$ in the model’s penultimate layer and $W=(w_1,...,w_C)$ represents the transformation coefficients of the last linear layer before the softmax layer. So the likelihood probability of data $x$ corresponding to the $M$ categories can be formulated as 
\begin{equation}
    \hat{y}=\mathcal{C}(x)=Softmax(W^T f_\theta(x)).
    \label{eqn:classification}
\end{equation}
Note that each vector $w_i$ in matrix $W$ can be considered as the prototype of class $i$ and the production $W^T f_\theta(x)$ in (\ref{eqn:classification}) quantifies the similarity between the feature $f_{\theta}(x)$ and different class-prototypes. We follow the theory in \cite{advFeature} and rephrase the definitions of \textit{robust} and \textit{non-robust} features for analysis in this paper.
\begin{itemize}
\item \textbf{\textit{$\rho-$}useful features} refer to the features $f_U(x)$ correlated with the true class prototype, $w_t\cdot f_U(x) \geq \rho$. A classifier $\mathcal{C}$ utilizing $f_U(x)$ for decisions should yield non-trivial performance. 

\item \textbf{Irrelevant features}, denoted by $f_I(x)$, introduce background noise across all classes and do not contribute to the prediction. $f_I(x)=f_\theta(x)-f_U(x)$.

\item \textbf{\textit{$\gamma-$}robust features} are defined as those \textit{$\rho-$} useful features with respect to $x$ that are still $\gamma-$useful with respect to $x+\delta$, i.e.
$w_t\cdot f_R(x+\delta) \geq \gamma$. Note that in order to estimate meaningful robust features, $\gamma$ should be close to $\rho$ so that $w_t\cdot f_R(x+\delta)\approx w_t\cdot f_R(x)$.

\item \textbf{Non-robust features} can be easily defined as those features that are useful but not robust to adversarial perturbations. That is, $ f_{NR}(x)=f_U(x)-f_R(x)$ and $w_t\cdot f_{NR}(x+\delta)\neq w_t\cdot f_{NR}(x)$. Classifiers that rely largely on $f_{NR}(x)$ will easily be fooled.
\end{itemize}
Thus, given a model $\mathcal{C}$, $f_\theta(x)$ can be decomposed as
\begin{equation}
    f_\theta(x)=f_{R}(x)+f_{NR}(x)+f_{I}(x).
    \label{eq:fd}
\end{equation}

\subsection{Logit Change Patterns in Adversarial Attack}
A \underline{\textbf{logit}} is defined as $\pi_i=w_i\cdot f_\theta(x)$, indicating how strong the feature $f_\theta(x)$ is correlated with class $i$ \citep{Concise}. In classification, $\pi_t>\pi_i$ for $i\neq t$ guarantees a correct prediction. When an adversarial attack succeeds, $\pi'_a>\pi'_t$ and $\pi'_a>\pi'_i$ for $i\neq a$. In this paper, the subscript $t$ and $a$ indicate the ground-truth category and the adversarial category, respectively.

To further understand the behavior of adversarial attacks, we analyze the difference in logits before and after an attack. Specifically, with the feature decomposition in (\ref{eq:fd}), the logit changes associated with the true class and adversarial class approximate 
\begin{equation}
\begin{split} \label{eq:logitDiff}
    \Delta \pi_t &=\pi'_t-\pi_t \approx w_t\cdot [f_{NR}(x+\delta)-f_{NR}(x)]+\varphi_t\\
    \Delta \pi_a &=\pi'_a-\pi_a \approx w_a\cdot [f_{NR}(x+\delta)-f_{NR}(x)]+\varphi_a,
\end{split}
\end{equation}
where $\varphi_t$ and $\varphi_a$ are small constants accounting for minor logit changes associated with $f_I(x)$. From (\ref{eq:logitDiff}), the logit change is highly related to $f_{NR}(x+\delta)-f_{NR}(x)$ and its degree of alignment with the target or adversarial prototype. 

For a \underline{successful} adversarial attack $C(x')\neq y$, $\pi_t > \pi_a$ and $\pi_t'<\pi_a'$; the corresponding change pattern of $(\Delta \pi_t, \Delta \pi_a)$ can be categorized into three cases:
\begin{enumerate}
    \item ($\Delta \pi_t <0, \Delta \pi_a >0$): This category of attacks permutes data $x$ toward the adversarial category. Consequently, the feature $f_{NR}(x+\delta)$ is highly aligned with the adversarial prototype $w_a$, which makes $w_a\cdot f_{NR}(x+\delta)$ very large, meanwhile decreases $w_t\cdot f_{NR}(x+\delta)$.
    \item ($\Delta \pi_t << 0,  \Delta \pi_a \leq 0$): In this category of attacks, the adversary tends to mask discriminate features of the true class. The similarity between $w_t$ and $f_{NR}(x+\delta)$ significantly decrease, while $w_a\cdot f_{NR}(x+\delta)$ has a slight decrease.
    \item ($\Delta \pi_t \geq 0, \Delta \pi_a >> 0$): This is a rare case of adversarial attack where $f_{NR}(x+\delta)$ is highly aligned with adversarial prototype $w_a$ that makes $w_a\cdot f_{NR}(x+\delta)$ very large, meanwhile $w_t\cdot f_{NR}(x+\delta)$ has a minor increase.
\end{enumerate}

Other combinations of $(\Delta \pi_t, \Delta \pi_a)$ associate with \underline{unsuccessful} attacks. In this paper, unsuccessful attacks refer to those adversarial samples that fail to fool a classifier, i.e. $C(x')=y$.

\noindent\textbf{Remark:} Among the 3 categories of successful attacks above, both case 1 and 2 experience a significant drop in the logit associated with the ground-truth class. This analysis is consistent with our empirical observations where 99.69\% of successful attacks on CIFAR-10 lead to the greatest logit decrease in their true categories. In addition, we also notice that the logits associated with categories aside from the real and adversarial classes only have marginal differences. We believe that these common patterns in successful attacks are mainly attributed to the greedy nature of adversary generation optimization. Prior arts have shown that in theory, adversarial attacks, especially those gradient-based attacks, are prone to inducing these logit change patterns \citep{Hou2023} \footnote{One can design an adversarial attack violating these common logit change patterns by applying regularization techniques. However, based on our experiments, compared to the common attack algorithms, these methods are usually with higher computational overhead, lower time efficiency, and a lower attack success rate.}.

\section{Adversarial Logit Update Against Adversaries}
Based on our discussion in Preliminaries, it is natural to question: given a successful adversarial attack and its clean counterpart, can we use the logit change patterns to infer the sample category? The answer is obviously Yes! In this regard, this section will first elaborate on the corresponding ALU principle to infer the ground-truth labels for successful attacks. Then we introduce a new paradigm for data classification based on ALU.
\subsection{ALU Principle}
Unlike the conventional way to decide the final label prediction for a model by taking \textit{argmax} across all logit values, our ALU principle proposes to make the decision for adversarial samples by comparing the logit before and after the attack.  Specifically, given an adversarial sample $x'$ and its clean counterpart $x$, we compute the logit vectors $\Pi'=\{\pi'_1,...,\pi'_M\}$ and $\Pi=\{\pi_1,...,\pi_M\}$ and their difference $\Delta \Pi=\Pi-\Pi'$. Then the likelihood corresponding to the $M$ categories is
\begin{equation}
    \hat{y}=Softmax(\Delta \Pi)=Softmax(\Pi-\Pi').
    \label{qn:ALU}
\end{equation}
That is, the ALU principle selects the term with the largest logit increase as the final prediction.

\noindent\textbf{Remark:} In theory, the ALU principle fails case 3 (i.e. $\Delta \pi_t \geq 0, \Delta \pi_a >> 0$) and all unsuccessful attacks. We will show in the next section that incorporating a purifier for adversarial noise removal and a detector for potential "successful" adversary identification in the conventional classification paradigm would address this limitation.

\begin{figure*}[t]
    \centering
    \includegraphics[width=1\textwidth]{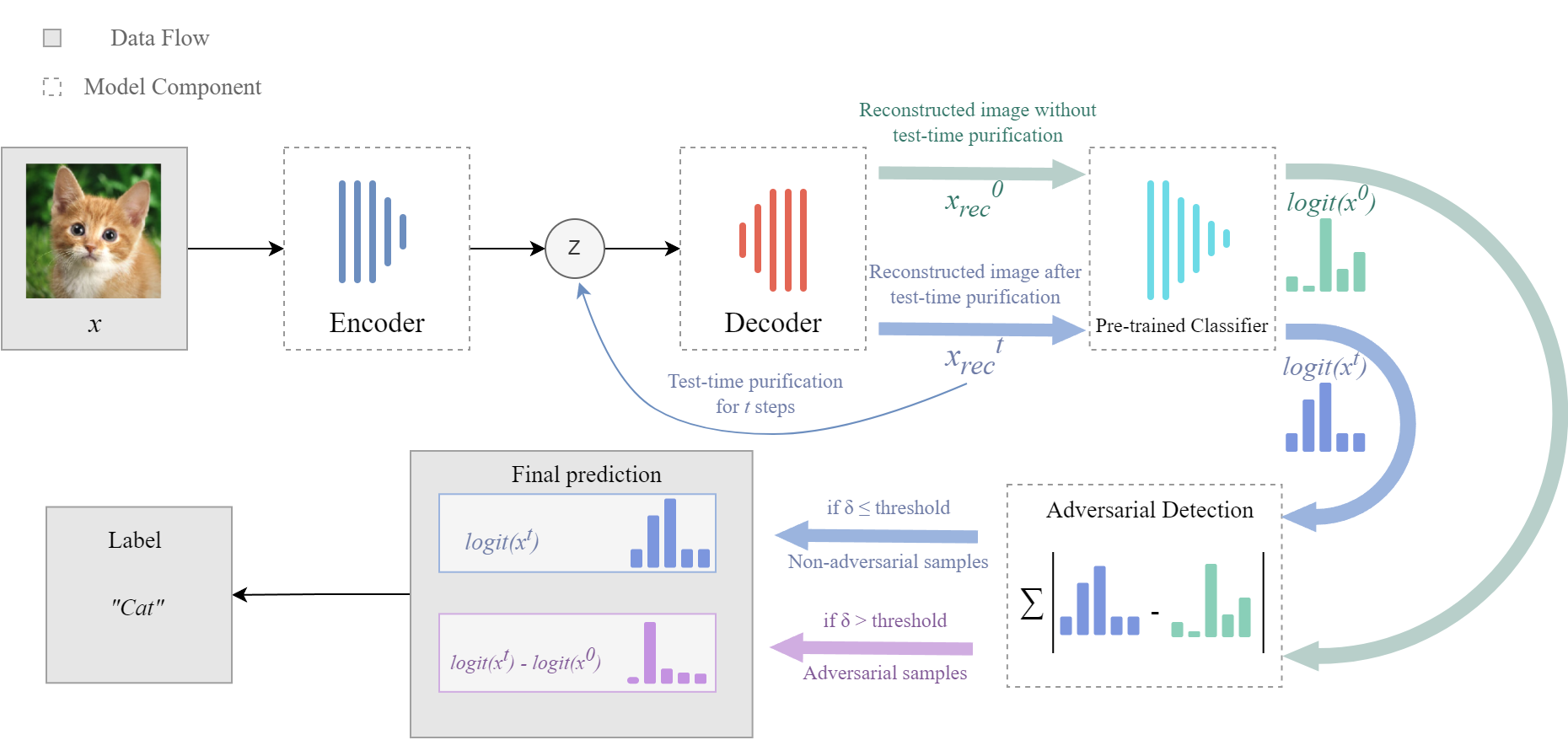}
    \caption{The diagram of our test-time ALU-based classification solution. Given a pre-trained model, a VAE is trained on clean data. In inference, the encoder of the VAE is dropped and the latent code $z$ is optimized to recover the input data.}
\label{Fig:main}
\end{figure*}

\subsection{Incorporating ALU in Classification Paradigm}
The ALU principle in (\ref{qn:ALU}) is complementary to the canonical prediction method in (\ref{eqn:classification}). Thus, combining the two classification principles within one paradigm would boost model adversarial performance. To this end, two issues need to be well addressed. 

First, applying the ALU principle to clean data would lead to incorrect predictions. So we need a good adversarial detection module to identify adversarial samples belonging to case 1 and 2 summarized in Preliminaries.

Second, applying the ALU principle requires to access an ideal clean version of $x'$, which is impossible in real applications. We hypothesize that purified data $\hat{x}$ can be used as a surrogate clean sample in ALU. To validate our hypothesis, we train a PGD-purification model on CIFAR-10 following \cite{Liao_2018_CVPR} and monitor the classification accuracy over various adversarial samples. As shown in Table \ref{main:unet}, incorporating ALU and purification model would significantly boost the model's adversarial robustness. The specific settings are specified in Section C of our Supplementary material. 
\begin{table}
    \centering
    \small
    \renewcommand{\arraystretch}{1.5}
    \caption{Adversarial robustness (\%) when incorporating a toy adversarial purification model (i.e. denoising U-Net \cite{Liao_2018_CVPR}) and the ALU principle on CIFAR-10.}
    \begin{tabular}{lcccc}
        \toprule
        Method& PGD20& C\&W($L_\infty$)& C\&W($L_2$)& AA\\
        \hline\hline
        Standard& 0& 3& 41& 0\\
        +purification& 18& 65& 57& 55\\
        +ALU & 70& 88& 80& 78\\
        \bottomrule
    \end{tabular}
    \label{main:unet}
\end{table}

In sum, the recipe for incorporating ALU into the conventional classification paradigm includes a data purifier, a pre-trained classifier, and an adversarial sample detector. The pseudo-code of the classification paradigm with the ALU principle is presented in Algorithm \ref{al:ALUCP}.

\begin{algorithm}[htbp]
\caption{Classification paradigm with the ALU principle}
\label{al:ALUCP}
\begin{algorithmic}[1]
\Procedure {Classification}{$x$} 
    \State Load a pre-trained classifier $\mathcal{C}=(f_{\theta},W)$
    \State Load an adversarial purifier $P$
    \State Load an adversarial sample detection $D$
    \State $\hat{x} \gets P(x)$ \Comment{Purified version of $x$}
    \State $\Pi' \gets W^T f_\theta(x)$ \Comment{Pre-purification logit}
    \State $\Pi \gets W^T f_\theta(\hat{x})$ \Comment{Post-purification logit}
    \If{$D(x)$} 
        \State $y \gets softmax(\Pi-\Pi')$  \Comment{ALU principle}
    \Else
        \State $y \gets softmax(\Pi)$  \Comment{Convential Classification}
    \EndIf
\EndProcedure
\end{algorithmic}
\end{algorithm}

\section{Test-time ALU for Model Robustness }
Algorithm \ref{al:ALUCP} provides a conceptual paradigm for data classification with the ALU principle. It provides a new perspective on model robustness enhancement. In this section, we present a simple, yet effective implementation based on Algorithm \ref{al:ALUCP}. The detailed systematic structure of our implementation is presented in Fig.\ref{Fig:main}, where a Variational Autoencoder (VAE) \cite{VAE2013} is used to generate purified data in an iterative manner for clean image synthesis.

\subsection{Test-time Purification for Clean Image Synthesis}
The proposed ALU principle needs to be paired with an adversarial purification model to work. We attempt to replace our toy purification model in Table 1 with more sophisticated methods for performance boost but fail. We hypothesize two reasons for this outcome. First, SOTA purification solutions usually require pairs of adversarial samples and their clean versions as training data \cite{Liao_2018_CVPR,ref9}. However, such purifiers targeting a specific attack are fragile against other attacks. Second, we notice that many purification models also require an adversarially-trained model as the backbone classifier, otherwise showing poor performance.

Since adversarial perturbation varies with the attacked model, attack method, and data, we aim to construct a model-agnostic purifier for clean data synthesis. To this end, VAE is taken as the backbone of the purifier and trained with clean images only. 
The continuity in VAE's latent space facilities the generation of good surrogate clean images.

In inference, given a query $x$, we initialize the latent code $z$ randomly or by $z=E(x)$, and optimize the initial $z$ to reconstruct input $x$ iteratively:
\begin{equation}
    \min_{z}(D(z),x)=||D(z)-x||_2,
\end{equation}
where $E$ stands for the encoder and $D$ represents the decoder.

Notably, since VAE is trained on clean data only, the test-time iterative update doesn't hurt the synthesis quality for a clean input. If the input is adversarial attacked, VAE would fail to synthesize adversarial permutations. However, the strong visual content in an image helps restore its latent code back to where its clean version would be.

Theoretically, for any initial code $z$, there always exists a learning rate that guarantees a correct classification in just one step test-time adaptation (whose mathematical proof is provided in Section A of our Supplementary material). However, in reality, due to the complex surface of the loss function and the lack of true labels in testing, a one-step update is impossible. Thus we propose to update the latent code $z$ by gradient descent with a small learning rate. Correspondingly, we use $x_{rec}^{t}$ to denote the synthesis output after $t$-step update. When $t=0$, the synthesis image $x_{rec}^0$ is generated directly from the adversarial latent code $E(x)$ without any update. With increasing update step $t$, $x_{rec}^t$ approaches its clean version.

\begin{figure}[t]
    \centering
    \includegraphics[width=0.4\textwidth]{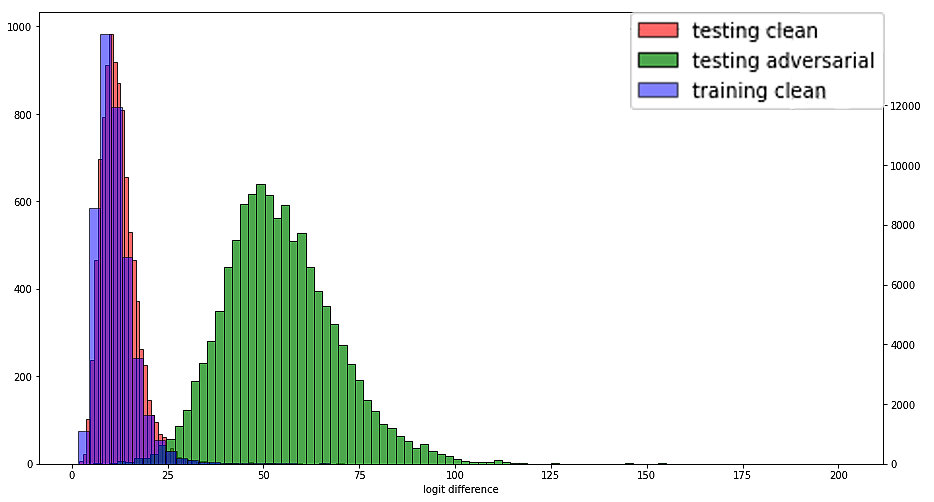}
    \caption{Histograms for logit difference before and after test-time update with CIFAR-10 training set and test set. The adversarial data is generated using PGD20.}
    \label{hist.logit_diff}
\end{figure}

\subsection{Integrated Statistical Adversarial Detection}
Though the ALU principle can effectively infer the true labels by comparing pre- and post-purification logits, it is not applicable to clean images. 

In this solution, we design a simple algorithm to distinguish adversarial samples from clean data using statistical information obtained in training. In particular, with the clean training data, we count the sum of logit value changes before and after purification, $sum(|\Delta \Pi|)$, and calculate a threshold that includes the majority of clean image statistics, 99.5\% in practice, to avoid extreme cases. 
During inference, if an input's logit change around purification is less than the threshold, we consider it clean data and use the post-purification logit to make predictions; otherwise, we will consider it an adversarial sample and use logit comparison for final output. Our empirical observation on $sum(|\Delta \Pi|)$ depicts in Fig.\ref{hist.logit_diff}, where statistical distributions of training and testing set significantly overlap, while those distributions of clean data and adversarial samples are highly separable.

\noindent\textbf{Remark:} The proposed method takes VAE, a fairly simple generative model, for "clean" data synthesis, and utilizes a detection solution that is also a bit ad-hoc. However, we believe that if such a design shows excellent performance, the classification paradigm with the ALU principle shown in Algorithm \ref{al:ALUCP} should work only better on other advanced generative models and detection algorithms.

\subsection{ALU Extension: Unified Classifier Training}

In our solution, VAE is used to synthesize clean samples. However, VAE's outputs usually are blurry. This may pose a challenge for a brittle classifier. To further improve model robustness, the VAE and classifier can also be jointly trained from scratch to boost both clean and adversarial accuracy. We call this ALU extension as \textbf{Unified ALU}. In Unified ALU, we first train the generative model, VAE in our case, using clean data, and then the classifier is trained using VAE's output. Note that despite achieving outstanding performance, our unified training classifier is also trained using the standard training procedure \textit{without adversarial data}.

\section{Experimentation and Discussions}

\subsection{Experimental Setup and Implementation Details}
We evaluate our solution on CIFAR10, CIFAR100, and tiny-ImageNet. Model robustness against PGD-20 and AutoAttack with an epsilon of 8/255 is reported. In evaluation, our ALU-based methods are attacked with gradient calculated through both generative model and classifier to make it a full White-box attack.

Our model consists of a VAE (whose specific architecture is depicted in Section B of our Supplementary material) for clean image synthesis and a ResNet50 for image classification. 
Both VAE and ResNet50 are trained using ADAM optimizer with clean data only. For VAE, the learning rate is set to be 0.001 with deduction rates of 0.2 and 0.04 at epoch 80 and 100. For ResNet50, we have a learning rate of 0.001 and deduction rates of 0.2 and 0.04 at epoch 100 and 120. All experiments run on one RTX3090.

\begin{table*}[htbp]
    \caption{Adversarial robustness (\%) when incorporating ALU with various pre-trained classifiers. Note, white-box attacks (i.e. adversary knows the detailed structures of both classifier and VAE) are used in this evaluation. Adversarially-trained models are marked with $\dagger$ in the table.}
     \small
    \centering
    \renewcommand{\arraystretch}{1.5}
    \begin{tabular}{cccccccccc}
    \toprule
    \multirow{2}[1]{*}{Method}& \multicolumn{3}{c}{CIFAR10}& \multicolumn{3}{c}{CIFAR100}& \multicolumn{3}{c}{tiny-ImageNet}\\
    \cmidrule(lr){2-4} \cmidrule(lr){5-7} \cmidrule(lr){8-10}
     & Clean& PGD20& AutoAttack& Clean& PGD20& AutoAttack& Clean& PGD20& AutoAttack\\
    \hline\hline
    \makecell[l]{Standard\\ +ALU}& \makecell[c]{\textbf{93.85}\\ 79.34}& \makecell[c]{0\\ 64.52}& \makecell[c]{0\\ 72.03}& \makecell[c]{\textbf{74.46}\\ 53.10}& \makecell[c]{0\\ 41.83}& \makecell[c]{0\\ 37.86}& \makecell[c]{\textbf{65.28}\\ 40.09}& \makecell[c]{0\\ 18.10}&  \makecell[c]{0\\ 21.21}\\
    \hline
    \makecell[l]{\cite{ref1}$\dagger$\\ +ALU}& \makecell[c]{88.83\\ 86.50}& \makecell[c]{48.68\\ \underline{96.87}}& \makecell[c]{45.83\\ 85.55}& \makecell[c]{62.07\\ 58.86}& \makecell[c]{23.64\\ 82.09}& \makecell[c]{22.29\\ 56.20}& \makecell[c]{49.20\\ 41.52}& \makecell[c]{13.62\\ 42.18}& \makecell[c]{12.20\\ \underline{35.97}}\\
    \hline
    \makecell[l]{\cite{ref2}$\dagger$\\ +ALU}& \makecell[c]{84.44\\ 82.22}& \makecell[c]{54.42\\ 94.18}& \makecell[c]{51.19\\ 81.71}& \makecell[c]{57.35\\ 54.34}& \makecell[c]{27.41\\ \textbf{87.89}}& \makecell[c]{25.15\\ 51.90}& \makecell[c]{45.25\\ 38.76}& \makecell[c]{19.15\\ \underline{55.98}}& \makecell[c]{15.30\\ 35.88}\\
    \hline
    \makecell[l]{\cite{ref3}$\dagger$\\ +ALU}& \makecell[c]{88.51\\ 86.07}& \makecell[c]{49.01\\ 96.83}& \makecell[c]{43.49\\ \underline{86.63}}& \makecell[c]{\underline{65.34}\\ 61.89}& \makecell[c]{23.04\\ 78.29}& \makecell[c]{21.46\\ \underline{58.65}}& \makecell[c]{48.16\\ 40.88}& \makecell[c]{12.68\\ 40.11}& \makecell[c]{11.34\\ 34.04}\\
    \hline
    \makecell[l]{\cite{ref27}$\dagger$\\ +ALU}& \makecell[c]{85.67\\ 81.93}& \makecell[c]{57.68\\ 79.13}& \makecell[c]{42.17\\ 66.16}& \makecell[c]{60.80\\ 57.50}& \makecell[c]{24.46\\ \underline{83.97}}& \makecell[c]{21.05\\ 51.64}& \makecell[c]{46.08\\ 36.48}& \makecell[c]{19.27\\ 49.32}& \makecell[c]{14.23\\ 30.60}\\
    \hline
    \makecell[l]{\cite{ref5}$\dagger$\\ +ALU}& \makecell[c]{83.52\\ 81.89}& \makecell[c]{57.62\\ 87.80}& \makecell[c]{51.12\\ 76.43}& \makecell[c]{58.61\\ 55.62}& \makecell[c]{33.84\\ 74.78}& \makecell[c]{28.52\\ 50.07}& \makecell[c]{39.26\\ 34.86}& \makecell[c]{21.69\\ 47.30}& \makecell[c]{16.41\\ 30.21}\\
    \hline
    Unified ALU & \multirow{1}[1]{*}{\underline{89.25}}& \multirow{1}[1]{*}{\textbf{97.97}}& \multirow{1}[1]{*}{\textbf{89.72}}& \multirow{1}[1]{*}{65.02}& \multirow{1}[1]{*}{81.35}& \multirow{1}[1]{*}{\textbf{59.54}}& \multirow{1}[1]{*}{\underline{50.43}}& \multirow{1}[1]{*}{\textbf{63.52}}& \multirow{1}[1]{*}{\textbf{49.24}}\\
    \bottomrule
    \end{tabular}
    \label{main_result1}
\end{table*}

\subsection{Effectiveness of ALU on pre-trained models }
We first apply the plug-in ALU solution on pre-trained models and report their performance in Table \ref{main_result1}. The prediction of various pre-trained models is based on the conventional maximum likelihood probability. For both vanilla- and adversarially-trained models, ALU boosts the adversarial robustness significantly, with a small degradation on standard accuracy. However, we notice that ALU harms the standard accuracy since the vanilla-trained model is brittle and not robust to VAE's blurring effect. We also include the Unified ALU in the table. Still using the vanilla training, unified ALU exposes the blurred data to the classifier before model deployment, thus leading to a classifier less sensitive to data purification.

\noindent\textbf{Remark:} An interesting observation is that our \textit{adversarial accuracy can even outpace clean accuracy}. This is theoretically impossible in the prior arts that predict labels based on absolute logits. 
We study carefully for this observation and discover that this surprising phenomenon is attributed to our novel ALU design. Following the ALU principle, the model can find the true label by comparing pre- and post-purification logits, even though the classification based on the absolute logits is not correct. For most samples that a classifier does wrong, they are considered hard samples that involve one or multiple classes as confusable classes. That is, the classifier partially recognizes the samples while outperformed by confusable classes. This means the true label logits are lower than confusable class logits while higher than all the other unrelated class logits. When under adversarial attacks, the true label logits will still drop significantly, which gives the opportunity for ALU to find true labels through logit comparison, providing higher adversarial accuracy than clean accuracy. 
\begin{table*}[htbp]
    \caption{Adversarial robustness (\%) compared to other test-time adaptation methods. Results collected from references are marked with *. Methods utilizing extra data (such as adversarial samples) are marked with $\dagger$.}
    \centering
    \small
    \renewcommand{\arraystretch}{1.5}
    \begin{tabular}{cccccccccc}
    \toprule
    \multirow{2}[1]{*}{Method}&  \multicolumn{3}{c}{CIFAR10}& \multicolumn{3}{c}{CIFAR100}& \multicolumn{3}{c}{tiny-ImageNet}\\
    \cmidrule(lr){2-4} \cmidrule(lr){5-7} \cmidrule(lr){8-10}
     & Clean& PGD20& AutoAttack& Clean& PGD20& AutoAttack& Clean& PGD20& AutoAttack\\
    \hline\hline
    \cite{ref28}$\dagger$& 81.88& 58.59& 59.44& 56.09& 34.37& 27.96& 46.97& 14.48& 19.33\\
    \hline
    \cite{ref32}*$\dagger$& 90.63& 71.43& \underline{72.66}& \underline{68.67}& \underline{43.25}& \underline{45.00} & - & - & -\\
    \hline
    \cite{ref7}*$\dagger$& 88.25& 63.73& 60.05& 54.39& 30.71& 28.86& - & - & -\\
    \hline
    \cite{ref23}*& \underline{91.89}& 53.58& -& 61.01& 37.53& -& - & - & -\\
    \hline
    \cite{ref37}*& \textbf{93.09}& \underline{85.45}& -& \textbf{77.83}& 43.21& -& - & - & -\\
    \hline
    \cite{ref38}& 84.80& 78.91*& 40.20& 51.40& 26.10*& 18.60& - & - & -\\
    \hline
    Unified ALU& 89.25& \textbf{97.97}& \textbf{89.72}& 65.02& \textbf{81.35}& \textbf{59.54}& \textbf{50.43}& \textbf{63.52}& \textbf{49.24}\\
    \bottomrule
    \end{tabular}
    \label{main_result2}
\end{table*}

\subsection{Comparison to test-time adaptation-based methods}
In this experiment, Table \ref{main_result2} compares our unified ALU method as a whole piece to existing methods that deploy test-time adaptation mechanisms to improve the model's adversarial robustness. Again, ALU achieves competitive standard accuracy but significantly strong adversarial robustness. This is because ALU tackles the problem from a new perspective which helps to make the right prediction on adversarial samples, even though their classifications based on the likelihood are not correct. 

\noindent\underline{\textbf{Limitation:}} As discussed in \cite{ref21}, test-time adaptation methods are vulnerable to BPDA-EoT \citep{athalye18a} attack which can attack through non-differentiable parts in test-time adaptation as well as randomized defenses. We examine closely in this attack and find that BPDA-EoT attacks the VAE model and makes it unable to find the right path back to the clean image. While this is a concerning problem for ALU connected with pre-trained classifiers, it can be easily avoided by modifying the unified ALU scheme. Since BPDA-EoT attacks through non-differentiable defenses by attacking $\hat{x}$ which is the generated (purified) image fed into the classifier, and applies the generated adversarial noise on the original image $x$, the attack will not work if the generative model has asymmetric input and output. That is, $x$ and $\hat{x}$ have different dimensions. In this regard, the generative model can be trained normally, however, instead of feeding the normal output into the classifier, other latent codes (output of inter-layers) of the model can be used as input to the classifier. Classifier structures need to be slightly changed accordingly.

\subsection{Ablation Study on Hyper-parameter Setting}
\begin{figure}[htbp]
    \begin{subfigure}[htbp]{0.25\textwidth}
        \begin{tikzpicture}[scale=0.48]
        \pgfplotsset{
            xlabel = {leaning rate (E-03)},
            ylabel = {accuracy},
            xmin = 0, xmax = 10,
            ymin = 0.7, ymax = 1.0,
            xtick = {0,1,2,3,4,5,6,7,8,9,10},
            ytick = {0.7,0.75,0.8,0.85,0.9,0.95,1.0},
            legend style={at={(1,1.11)},anchor=east},
            xmajorgrids=true,
            ymajorgrids=true,
        }
        \begin{axis}
        \addplot[
            color=blue,
            mark=o
            ]
            coordinates {(0.1,0.8966)(1,0.8943)(2,0.8905)(3,0.8783)(4,0.8647)(5,0.8453)(6,0.8235)(7,0.801)(8,0.7806)(9,0.7604)(10,0.7496)};
            \addlegendentry{clean accuracy}
        \addplot[
            color=red,
            mark=square
            ]
            coordinates {(0.1,0.9831)(1,0.9829)(2,0.9803)(3,0.9681)(4,0.9524)(5,0.933)(6,0.9106)(7,0.8957)(8,0.8786)(9,0.8622)(10,0.8496)};
            \addlegendentry{adversarial accuracy}
        \end{axis}
        \end{tikzpicture}
    \end{subfigure}%
    \begin{subfigure}[htbp]{0.25\textwidth}
        \begin{tikzpicture}[scale=0.48]
        \pgfplotsset{
            xlabel = {leaning rate (E-03)},
            xmin = 0, xmax = 10,
            xtick = {0,1,2,3,4,5,6,7,8,9,10},
            legend style={at={(1,1.11)},anchor=east},
            xmajorgrids=true,
            ymajorgrids=true,
        }
        \begin{axis}[
            ylabel = {adversarial detection rate},
            axis y line*=left,
            ymin = 0, ymax = 1.0,
            ytick = {0,0.1,0.2,0.3,0.4,0.5,0.6,0.7,0.8,0.9,1},
        ]
        \addplot[
            color=red,
            mark=square
        ]
        coordinates {(0.1,0.9945)(1,0.9815)(2,0.6857)(3,0.1823)(4,0.0547)(5,0.0463)(6,0.0639)(7,0.0629)(8,0.0749)(9,0.0892)(10,0.0592)};
        \label{lrdr_adv}
        \end{axis}
        \begin{axis}[
            ylabel = {clean detection rate},
            axis y line*=right,
            axis x line=none,
            ymin = 0.95, ymax = 1,
            ytick = {0.95,0.96,0.97,0.98,0.99,1},
        ]
        \addplot[
            color=blue,
            mark=o
        ]
        coordinates {(0.1,0.9892)(1,0.9916)(2,0.9911)(3,0.9862)(4,0.985)(5,0.985)(6,0.9844)(7,0.985)(8,0.9847)(9,0.9843)(10,0.9861)};
        \addlegendentry{clean detection rate}
        \addlegendimage{/pgfplots/refstyle=lrdr_adv}\addlegendentry{adversarial detection rate}
        \end{axis}
        \end{tikzpicture}
    \end{subfigure}
    \caption{Ablation on learning rate with CIFAR10 dataset. Left: learning rates versus accuracy. Right: learning rates versus clean and adversarial detection rate. The iteration number is fixed at 100 in this experiment, and we assume 100\% detection rate when measuring accuracy. Note that the learning rate starts from 0.1 instead of 0.}
    \label{Fig:lr}
\end{figure}
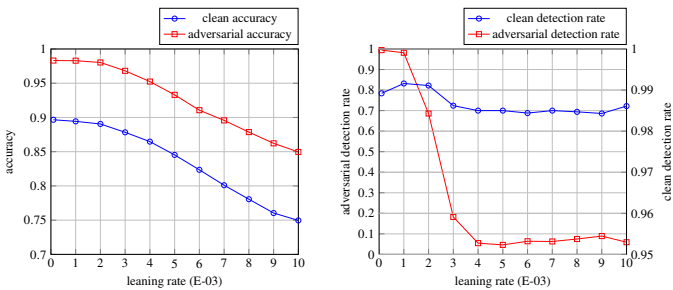

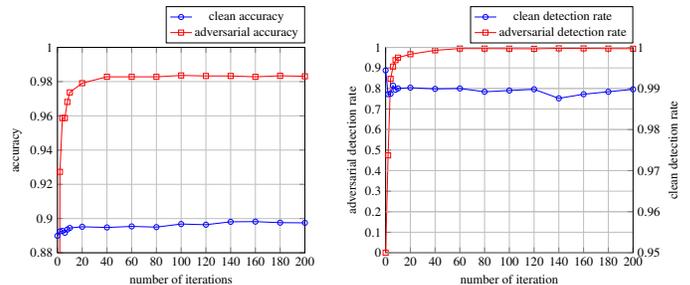
\begin{figure}[htbp]
    \begin{subfigure}[htbp]{0.25\textwidth}
        \begin{tikzpicture}[scale=0.48]
        \pgfplotsset{
            xlabel = {number of iterations},
            ylabel = {accuracy},
            xmin = 0, xmax = 200,
            ymin = 0.88, ymax = 1.0,
            xtick = {0,20,40,60,80,100,120,140,160,180,200},
            ytick = {0.88,0.9,0.92,0.94,0.96,0.98,1},
            legend style={at={(1,1.11)},anchor=east},
            xmajorgrids=true,
            ymajorgrids=true,
        }
        \begin{axis}
        \addplot[
            color=blue,
            mark=o
            ]
            coordinates {(0,0.8899)(2,0.8925)(4,0.8928)(6,0.8917)(8,0.8936)(10,0.8945)(20,0.8952)(40,0.8948)(60,0.8954)(80,0.895)(100,0.8968)(120,0.8964)(140,0.8981)(160,0.8982)(180,0.8976)(200,0.8975)};
            \addlegendentry{clean accuracy}
        \addplot[
            color=red,
            mark=square
            ]
            coordinates {(0,0.0863)(2,0.9273)(4,0.9587)(6,0.9587)(8,0.9681)(10,0.9737)(20,0.9791)(40,0.9828)(60,0.9828)(80,0.9828)(100,0.9836)(120,0.9833)(140,0.9833)(160,0.9828)(180,0.9834)(200,0.9831)};
            \addlegendentry{adversarial accuracy}
        \end{axis}
        \end{tikzpicture}
    \end{subfigure}%
    \begin{subfigure}[htbp]{0.25\textwidth}
        \begin{tikzpicture}[scale=0.48]
        \pgfplotsset{
            xlabel = {number of iteration},
            xmin = 0, xmax = 200,
            xtick = {0,20,40,60,80,100,120,140,160,180,200},
            legend style={at={(1,1.11)},anchor=east},
            xmajorgrids=true,
            ymajorgrids=true,
        }
        \begin{axis}[
            ylabel = {adversarial detection rate},
            axis y line*=left,
            ymin = 0, ymax = 1.0,
            ytick = {0,0.1,0.2,0.3,0.4,0.5,0.6,0.7,0.8,0.9,1},
        ]
        \addplot[
            color=red,
            mark=square
        ]
        coordinates {(0,0.0005)(2,0.4744)(4,0.8468)(6,0.9066)(8,0.9380)(10,0.9501)(20,0.9667)(40,0.9857)(60,0.9944)(80,0.9942)(100,0.9938)(120,0.9931)(140,0.9954)(160,0.9951)(180,0.9942)(200,0.9937)};
        \label{nidr_adv}
        \end{axis}
        \begin{axis}[
            ylabel = {clean detection rate},
            axis y line*=right,
            axis x line=none,
            ymin = 0.95, ymax = 1,
            ytick = {0.95,0.96,0.97,0.98,0.99,1},
        ]
        \addplot[
            color=blue,
            mark=o
        ]
        coordinates {(0,0.9944)(2,0.9886)(4,0.9887)(6,0.9907)(8,0.9897)(10,0.99)(20,0.9902)(40,0.9899)(60,0.99)(80,0.9892)(100,0.9895)(120,0.9898)(140,0.9876)(160,0.9886)(180,0.9892)(200,0.9898)};
        \addlegendentry{clean detection rate}
        \addlegendimage{/pgfplots/refstyle=nidr_adv}\addlegendentry{adversarial detection rate}
        \end{axis}
        \end{tikzpicture}
    \end{subfigure}
    \caption{Ablation on test-time iteration number with CIFAR10 dataset. Left: iteration numbers versus accuracy. Right: iteration numbers versus clean and adversarial detection rate. The learning rate in this experiment is fixed at 1E-04.}
    \label{Fig:ni}
\end{figure}

Our test-time update strategy needs two hyperparameters: the learning rate $\alpha$ and the iteration number $N$. We perform ablation experiments on both parameters to evaluate their influence on the final performance. All experiments are done on CIFAR10 using the same clean-trained ALU model. Adversarial data is generated using PGD20 white-box attack. Note that since our system uses two different ways to get the final label, the final result needs both absolute accuracy and detection rate high to be good. Fig. \ref{Fig:lr} shows ALU's performance versus various learning rates.
We observe that smaller learning rates are prone to lead to better performance. ALU's performance versus various iteration steps is summarized in Fig. \ref{Fig:ni}. Apart from some bad robustness on very low iteration numbers, both clean and adversarial accuracy tend to slightly increase with the iteration number and converge at very high iteration numbers.

\noindent\underline{\textbf{Limitation:}} Similar to many test-time adversarial defense strategies, the test-time procedure increases inference time. Running evaluations with 10000 CIFAR10 samples on one RTX3090, ALU takes 6.9 seconds for prediction when the iteration number is 0 (no test-time adaptation). With 200 iterations in the test-time adaptation process, it takes 166 seconds or approximately 24 times more. However, according to our ablation study in Fig. 7, adversarial accuracy will quickly converge as the iteration number increases and reach decent adversarial accuracy and detection rate around 50 iterations. Therefore, if time efficiency is considered a priority, one can use 50 iterations which will take 46.9 seconds or 6.8 times without test-time adaptation. 

\section{Conclusions}
In this paper, we examined the patterns of logit changes in adversarial attacks and proposed a novel principle, ALU, for DNN adversarial robustness enhancement. By comparing the differences in pre- and post-purification logits, ALU could accurately infer true labels with high confidence. Extensive experiments demonstrated that our method significantly outperformed prior approaches and achieves new SOTA. The new paradigm offered several advantages, such as: (1) it did not require adversarial or additional data for model training, (2) the ALU-based model was easy to train and had fewer hyperparameters, and (3) the ALU principle could be easily applied to pre-trained models to significantly improve robustness.

\bibliography{main}

\appendix

\onecolumn

This is the supplementary document of our paper, entitled "Advancing Adversarial Robustness Through Adversarial Logit Update". The Python code of the proposed solution is also submitted together with this document.

\section{Mathematical Proof}
In this section, we provide rigorous mathematical proof that \textbf{for any query data (either adversarial or clean), ideally, our test-time adaptation strategy is able to generate a "good" clean version from any latent code by one-step update} for downstream adversarial logit update (ALU). Here, the "good" means the prediction label of the generated clean data is correct.

\textbf{Notations:} Let $\boldsymbol{E}$ and $\boldsymbol{D}$ represent the encoder and decoder of a generative model, i.e. a VAE model in our solution. The downstream classification model is denoted as $\boldsymbol{C}$. For an clean image $x$, $\hat{x}$ is the corresponding synthetic sample generated by the generative model. $z$ is a randomly-initialized latent code for the generative model.

\textbf{Proof:} Following convention, the generative model is optimized using the MSE loss for data reconstruction:
\begin{equation}
    L_{rec}=(x-\hat{x})^{2}.
\end{equation}

$\hat{x}$ is generated using any latent code $z$. So we have
\begin{equation}
L_{rec}=(x-\boldsymbol{D}{\cdot}z)^{2}.
\end{equation}
To achieve $\hat{x}=x$, we perform the test-time learning, where we update the latent code $z$ using the gradient of the reconstruction loss. In this regard, we take the derivative of reconstruction loss w.r.t the latent code $z$ and have
\begin{equation}
\frac{d}{dz}L_{rec}=\frac{d}{dz}(x-\boldsymbol{D}{\cdot}z)^{2}
=2(x-\boldsymbol{D}{\cdot}z)\cdot(-\boldsymbol{D}^{T}).
\end{equation}
Here, we have our updated latent code $z'$ calculated as
\begin{equation}
z'=z-\alpha\frac{d}{dz}L_{rec}\\
=z-\alpha\cdot2(x-\boldsymbol{D}{\cdot}z)\cdot(-\boldsymbol{D}^{T})\\
=z+\alpha\cdot2(x-\boldsymbol{D}{\cdot}z)\cdot\boldsymbol{D}^{T},
\end{equation}
where $\alpha$ represents the learning rate for the update. 

For simplification, we first take MSE of the classification likelihood as the target loss to train the classifier. Later, we will show that the conclusion also hold with cross-entropy loss. The MSE between the predicted likelihood and the true label is
\begin{equation}
L_{MSE}=(y_0-y)^{2},
\end{equation}
where $y_0$ stands for the true label and $y$ is the prediction generatd by the classifier. Since $y=f(\hat{x};\boldsymbol{C})=\boldsymbol{C}\cdot\hat{x}$ and $\hat{x}=\boldsymbol{D}\cdot z'$,
\begin{equation}
L_{MSE}=(y_0-\boldsymbol{C}\cdot\boldsymbol{D}\cdot z')^2\\
=(y_0-\boldsymbol{C}\cdot\boldsymbol{D}\cdot(z+\alpha\cdot2(x-\boldsymbol{D}\cdot z)\cdot\boldsymbol{D}^T))^2.
\end{equation}
Therefore, our problem transforms into a minimization problem: find a learn rate $\alpha$ so that the prediction is correct.
\begin{equation}
\min\limits_\alpha(y_0-\boldsymbol{C}\cdot\boldsymbol{D}\cdot(z+\alpha\cdot2(x-\boldsymbol{D}\cdot z)\cdot\boldsymbol{D}^T))^2.
\end{equation}
We can easily derive that there exists a learning rate 
\begin{equation}
\alpha=\frac{y_0-\boldsymbol{C}\cdot\boldsymbol{D}\cdot z}{2\boldsymbol{C}\cdot\boldsymbol{D}(x-\boldsymbol{D}\cdot z)\cdot\boldsymbol{D}^T}
\end{equation}
that the classification loss will be eliminated after one step of updating.

\textbf{Remark 1:} It should be noted that the optimal learning rate in either (8) or (11) is impossible to get directly since we do not have information on $y_0$ in the actual test. So this the proposed solution, we propose the iterative update strategy with a small learning rate for latent code update. Our ablation study in the main manuscript shows that with increasing update step, the reconstruction by the generative model is converging towards the true clean data.

\textbf{Remark 2:} In order to update the latent code along the right direction, we need to make sure the learning rate is positive. To ensure $\alpha>0$, we only need to prove $\frac{y_0-\boldsymbol{C}\cdot\boldsymbol{D}\cdot z}{x-\boldsymbol{D}\cdot z}>0$. To this end, we need two assumptions.

\underline{Assumption 1:}  For a normal generative model that is trained using reconstruction loss, we assume $x$, the original image, is the best sample representing the given feature.

\underline{Assumption 2:}  For a well-trained classifier, its output giving a clean image will be better than the output of a noisy input.

Under these two assumptions, we have 
\begin{equation}
L(\hat{x},x)>0,  \text{and   }
L(f(\hat{x}),y_0)>L(f(x),y_0)>0
\end{equation}
Since $\hat{x}=\boldsymbol{D}*z$ and $f(\hat{x})=\boldsymbol{C}\cdot\boldsymbol{D}\cdot z$, we can safe to say that
\begin{equation}
sign(y_0-\boldsymbol{C}\cdot\boldsymbol{D}\cdot z)=sign(x-\boldsymbol{D}\cdot z).
\end{equation}
Since neither of those two terms would equal zero, we can prove that the optimal learning rate $\alpha$ is always positive.

\textbf{Remark 3:} We can also prove that there is a learning rate $\alpha$ for one-step update toward clean data generation using Cross Entropy (CE) Loss as the classification optimization target. The Cross Entropy Loss is defined as
\begin{equation}
L_{CE}=-\sum_{i=1}^{C}p_i\cdot log(q_i),
\end{equation}
where $p_i$ is the true distribution and $q_i$ is the predicted distribution. In the canonical classification setting, the true distribution is often represented in the one-hot code form. So the CE loss can be written as
\begin{equation}
L_{CE}=-y_0\cdot log(y)\\
=-y_0\cdot log(\boldsymbol{C}\cdot\boldsymbol{D}\cdot z')\\
=-y_0\cdot log\left(\boldsymbol{C}\cdot\boldsymbol{D}\cdot(z+\alpha\cdot2(x-\boldsymbol{D}\cdot z)\cdot\boldsymbol{D}^T)\right).
\end{equation}
By easy math, we can also obtain a learning rate
\begin{equation}
\alpha=\frac{\frac{1}{\boldsymbol{C}\cdot\boldsymbol{D}}-z\cdot y_0}{2(x-\boldsymbol{D}\cdot z\cdot y_0)\cdot\boldsymbol{D}^T}
\end{equation}
that can eliminate the classification loss after one step of updating. The rest is the same as the proof with MSE loss in classification.


\section{Architecture of VAE}
\begin{figure*}[htbp]
\centering
\includegraphics[width=1\textwidth]{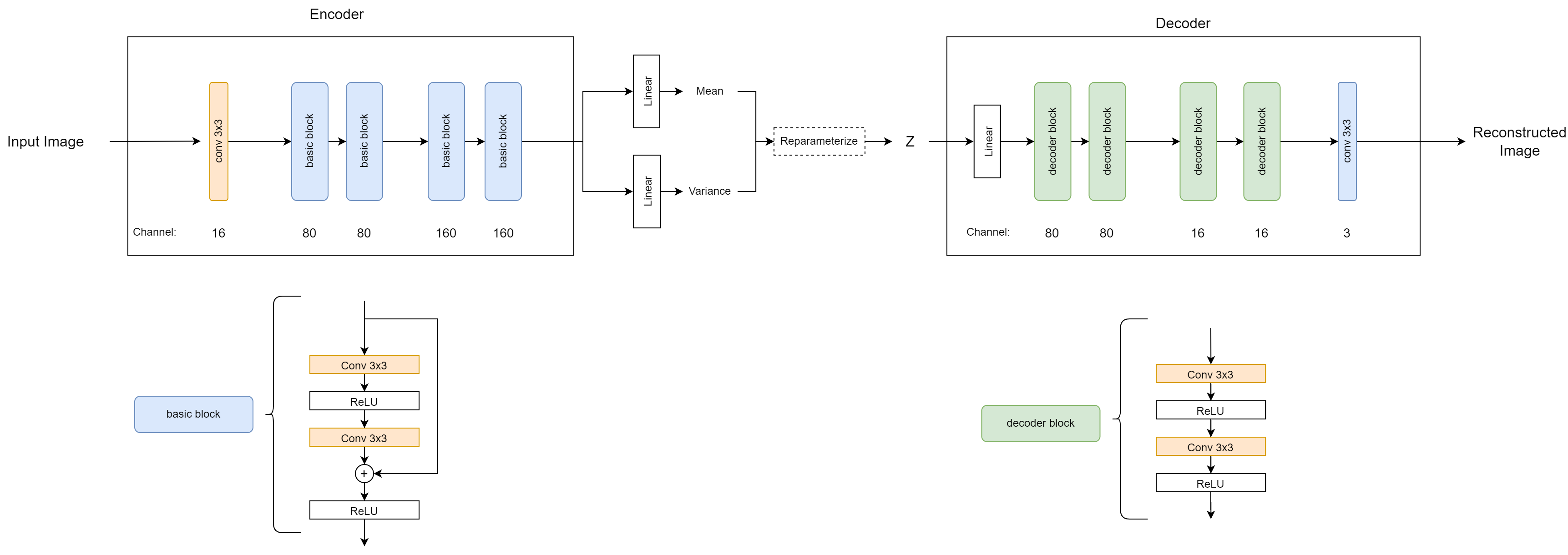}
\caption{Specific architecture of the VAE model used in our experiments.}
\label{Fig:VAE}
\end{figure*}

In the proposed solution, a VAE is exploited for surrogate "clean" data synthesis. Fig. \ref{Fig:VAE} shows the specific architecture of the VAE model used in our experiments.

\section{Details of Logit Comparison Toy Example for Table 1}
As described in the main paper, the ALU principle requires a surrogate task to find the "clean" version of adversarial samples. In the toy experiment presented in Table 1, we train a denoising U-Net following the scheme proposed in \citep{Liao_2018_CVPR}. Specifically, the adversarial noises used during U-Net training process are generated using a standard (non-robust) model with PGD20 attack. SGD optimizer is used with a learning rate of 0.1 and the epoch number is set to 100. All models are trained and tested on Cifar10 dataset. More details can be found in \citep{Liao_2018_CVPR}.

\section{Visualization on the ALU Principle}
\begin{figure*}[htbp]
\centering
\includegraphics[width=1\textwidth]{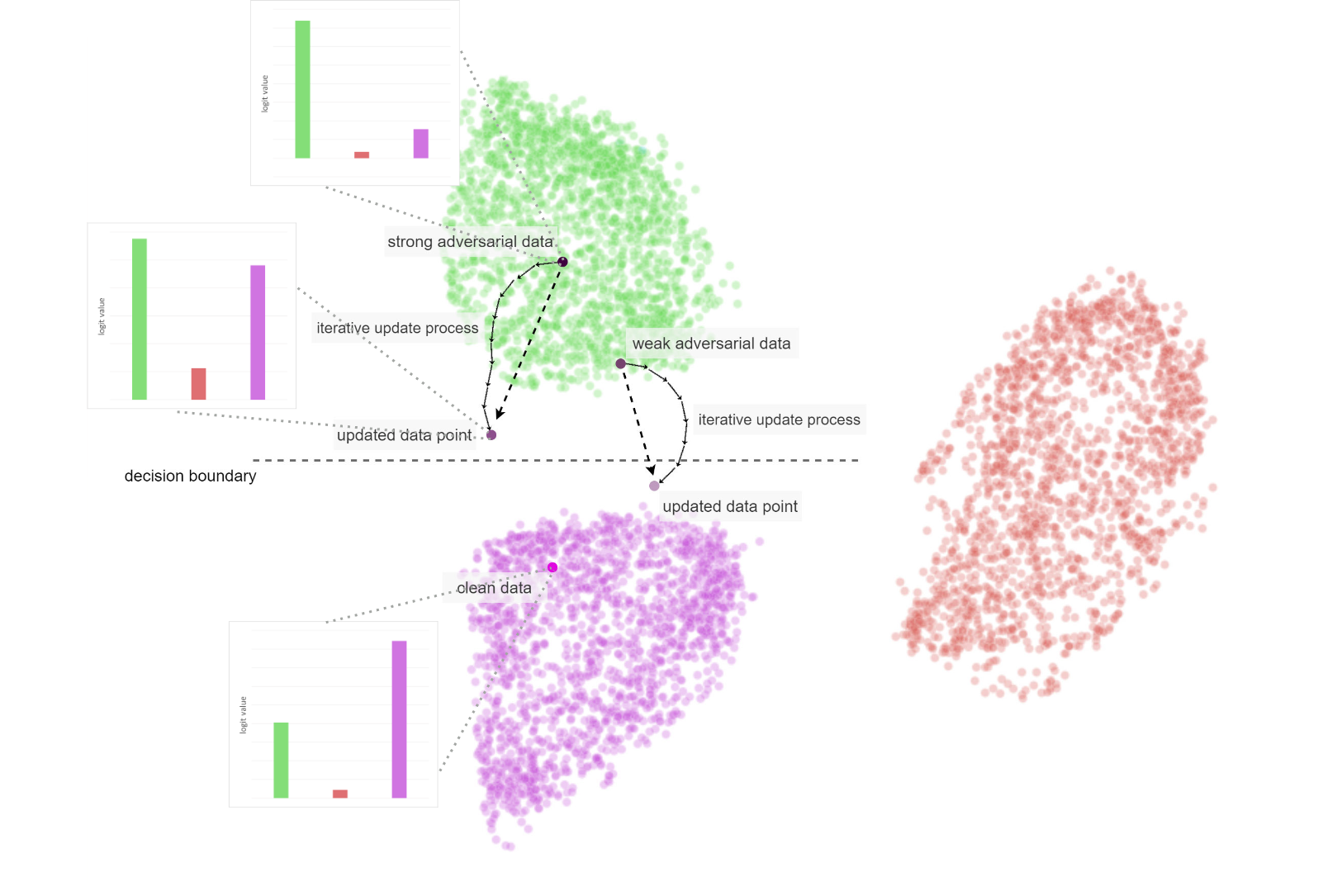}
\caption{Visualization of the ALU principle for adversarial robustness.}
\label{Fig:cluster}
\end{figure*}

Fig.\ref{Fig:cluster} provides a conceptual visualization of the ALU principle and the proposed solution. The column charts demonstrate the logit values corresponding to different data. It is noteworthy that though the estimated logit from the updated data may not equal the true value for strong adversarial samples, our ALU principle still works as long as $\hat{\Pi}-\Pi'$ follows the same or similar direction of $\Pi-\Pi'$. This also explains why the adversarial robustness can be higher than the standard classification accuracy when incorporating our ALU principle in the canonical classification paradigm.

\section{Empirical Observation on Logit Change}
In this section, we show the empirical observations on the logit changes around the adversarial attack that proves our theory about logit change patterns discussed in the main paper. 

\begin{figure*}[htbp]
\centering
    \begin{subfigure}[b]{0.45\textwidth}
        \includegraphics[width=1\linewidth]{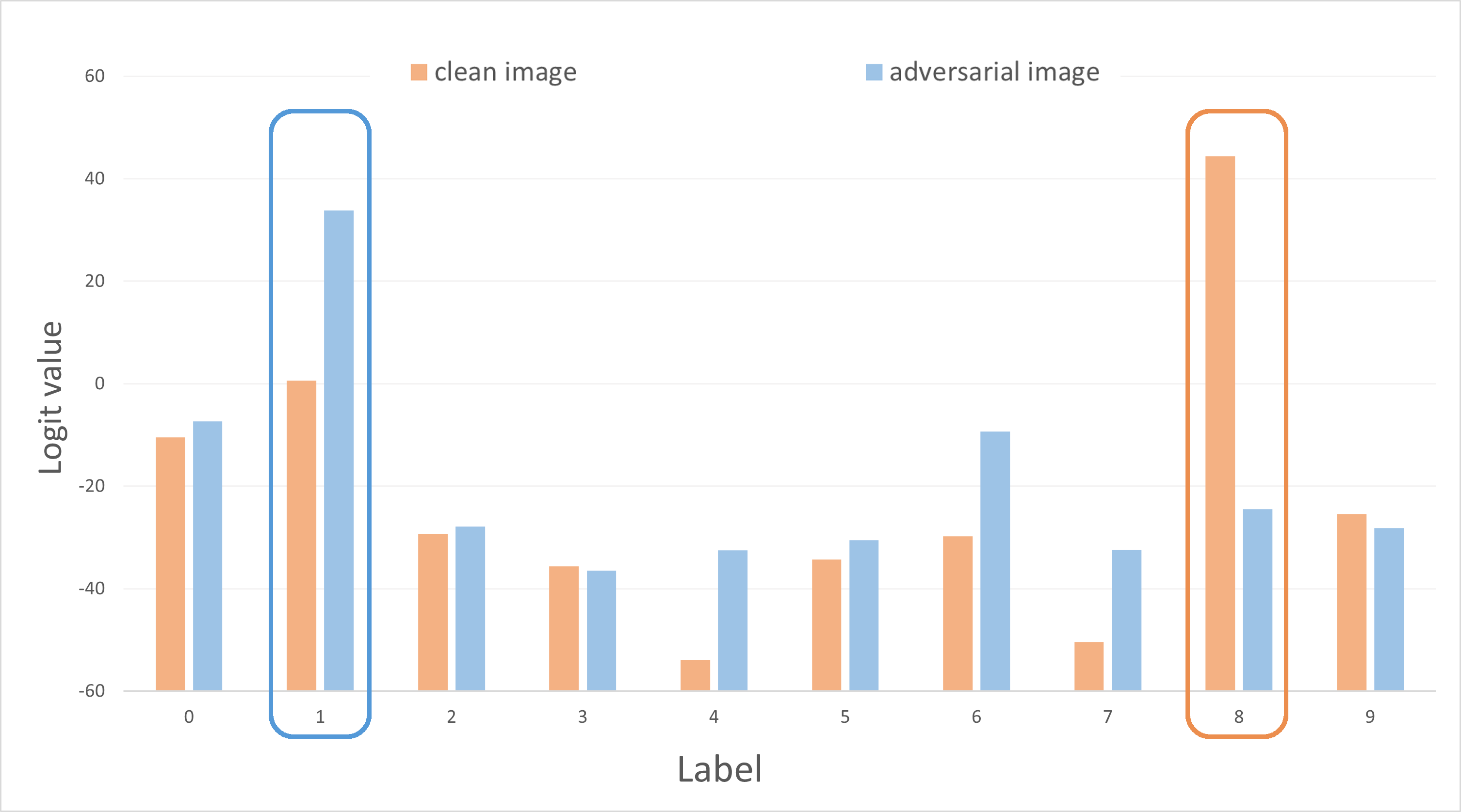}
    \end{subfigure}
    \hspace{0.7cm}
    \begin{subfigure}[b]{0.45\textwidth}
        \includegraphics[width=1\linewidth]{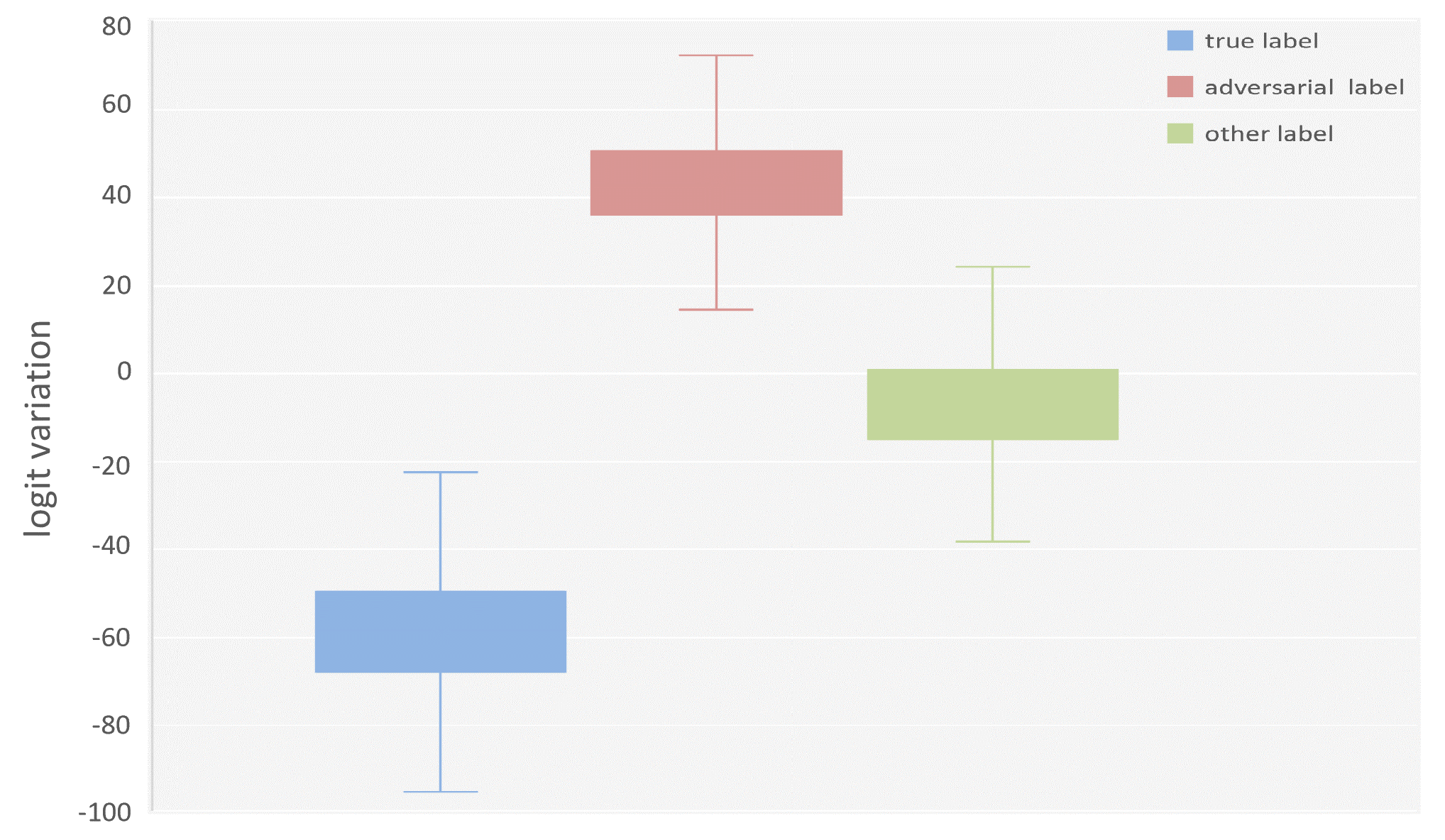}
     \end{subfigure}   
  \caption{Left: Example of logits change for a ship image (from the $8^{th}$ class in CIFAR-10) after the adversarial attack. Right: Statistics of logit changes, $\Pi'-\Pi$, on the CIFAR-10 test set. 99.69\% logit values of the true labels are significantly suppressed, while logit values of adversarial labels increase remarkably. Logit values associated with other classes show small perturbations after attacks.}
  \label {fig:logit_change}
\end{figure*}

The left plot of Fig.\ref{fig:logit_change} provides a representative example of the logit change around an adversarial attack. The orignal data is a Ship image (from the $8^{th}$ class) in CIFAR-10 dataset. After PDG-20 attack, the adversarial logit associated with the $2^{nd}$ class increases while the logit of the true label ($8^{th}$ category) is significantly suppressed. In addition, we monitor the logit changes in terms of $\Pi' - \Pi$, where $\Pi$ is the logit before perturbation and $\Pi'$ is the logit after perturbation, on CIFAR-10 test set and summarize the statistics on the right of Fig.\ref{fig:logit_change}. Compared to the increasing adversarial logit values and small perturbations in the logits associated with the other classes, the majority of the samples show a great or mostly greatest decrease in their true categories, which matches our discussion about logit change patterns in adversarial attack in the main paper.

\textbf{Remark:} As we stated in the main manuscript, these common patterns in successful adversarial attacks are mainly attributed to the greedy nature of adversary generation optimization. Though one can design an adversarial attack violating these common logit change patterns by applying regularization techniques. However, based on our experiments, compared to the common attack algorithms, these methods are usually with higher computational overhead, lower time efficiency, and a lower attack success rate.

\end{document}